\documentclass{vgtc}                          

\ifpdf
  \pdfoutput=1\relax                   
  \usepackage{graphicx}                
  \DeclareGraphicsExtensions{.pdf,.png,.jpg,.jpeg} 
\else
  \usepackage{graphicx}                
  \DeclareGraphicsExtensions{.eps}     
\fi%

\graphicspath{{figures/}{pictures/}{images/}{./}} 

\usepackage{microtype}                 
\PassOptionsToPackage{warn}{textcomp}  
\usepackage{textcomp}                  
\usepackage{mathptmx}                  
\usepackage{times}                     
\usepackage{cite}                      
\usepackage{tabu}                      
\usepackage{booktabs}                  

\usepackage{stfloats}
\usepackage[unskipbreak]{cuted}
\usepackage{lipsum}
\usepackage{graphicx, caption}
\input{insbox}
\usepackage{multirow}
\usepackage{subfigure}
\usepackage{enumitem}

\usepackage[hyphens]{url}
\usepackage{hyperref}
\hypersetup{colorlinks=true,breaklinks=true}

\usepackage{tikz}

\onlineid{7912}

\vgtccategory{Machine learning & explainable AI for cybersecurity}

\vgtcinsertpkg

\newlist{Req}{enumerate}{1}
\setlist[Req]{label=\textbf{R\arabic*:}}

\newlist{Task}{enumerate}{1}
\setlist[Task]{label=\textbf{T\arabic*:}}

\title{Visualization Of Class Activation Maps To Explain AI Classification Of Network Packet Captures}

 \author{Igor Cherepanov\thanks{e-mail: igor.cherepanov@igd.fraunhofer.de}\\ %
 \and Alex Ulmer\thanks{e-mail: alex.ulmer@igd.fraunhofer.de}\\ %
 \and Jonathan Geraldi Joewono \thanks{e-mail: jonathan.geraldi.joewono@igd.fraunhofer.de}\\ %
 \and J\"orn Kohlhammer\thanks{e-mail: joern.kohlhammer@igd.fraunhofer.de}\\ %
 }
 \affiliation{\scriptsize Fraunhofer IGD \\ Technische Universit\"at Darmstadt, Germany}

\teaser{
  \centering
  \includegraphics[width=\linewidth]{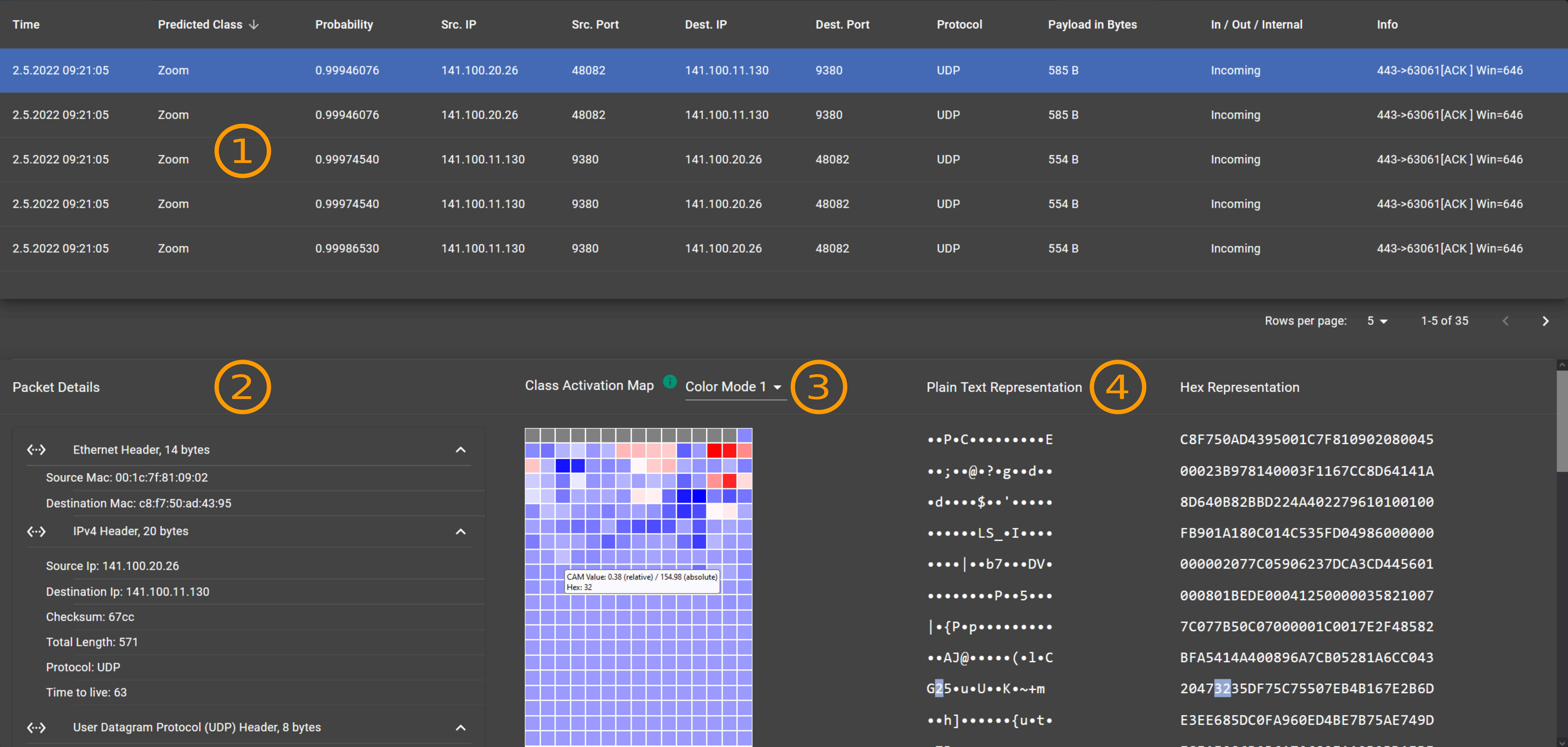}
  \caption{An overview of our classification system for the PCAP data. The system classifies the PCAPs and provides an explanation of the classifications based on the Class Activation Map technique (CAM).
  In the upper part (1) is a list of the packets that have been classified by our model. The columns of the tables are based on the extracted information of the PCAP files and two of them such as probability and predicted class are returned by the model.
  The packet details view (2) represents extracted information from the protocol headers.
  The CAM (3) represents the explanation of the corresponding PCAP classification. 
  Each grid in the CAM represents a byte of the selected packet.
  The colors illustrate the impacting bytes that led to this class prediction, ranging from \textit{blue} representing no relevance to \textit{red} representing maximum relevance.
  The selected packet is illustrated additionally in plain text and hex representation (4).}
  \label{fig:teaser}
}

\abstract{
The classification of internet traffic has become increasingly important due to the rapid growth of today's networks and application variety.
The number of connections and the addition of new applications in our networks causes a vast amount of log data and complicates the search for common patterns by experts.
Finding such patterns among specific classes of applications is necessary to fulfill various requirements in network analytics.
Supervised deep learning methods learn features from raw data and achieve high accuracy in classification.
However, these methods are very complex and are used as black-box models, which weakens the experts' trust in these classifications.
Moreover, by using them as a black-box, new knowledge cannot be obtained from the model predictions despite their excellent performance.
Therefore, the explainability of the classifications is crucial. 
Besides increasing trust, the explanation can be used for model evaluation to gain new insights from the data and to improve the model.
In this paper, we present a visual and interactive tool that combines the classification of network data with an explanation technique to form an interface between experts, algorithms, and data. 
} 

\newcommand\copyrighttextmy{%
  \footnotesize \textcopyright 2022 IEEE. Personal use of this material is permitted. Permission from IEEE must be obtained for all other uses, in any current or future media, including reprinting/republishing this material for advertising or promotional purposes, creating new collective works, for resale or redistribution to servers or lists, or reuse of any copyrighted component of this work in other works.
  DOI: \href{https://ieeexplore.ieee.org/Xplore/home.jsp}{10.1109/VizSec56996.2022.9941392}
  }
\newcommand\copyrightnotice{%
\begin{tikzpicture}[remember picture,overlay]
\node[anchor=south,yshift=10pt] at (current page.south) {\fbox{\parbox{\dimexpr\textwidth-\fboxsep-\fboxrule\relax}{\copyrighttextmy}}};
\end{tikzpicture}%
}

\CCScatlist{
    \CCScatTwelve{Human-centered computing}{Visu\-al\-iza\-tion}{User interface design}
    \CCScatTwelve{Interpretability}{Network Classification}{Convolutional Neural Networks}
}

\begin{document}



\firstsection{Introduction}

\maketitle
\copyrightnotice
The constant growth of connections and appearance of new applications in our networks remains Network Traffic Monitoring and Analysis (NTMA) as an important research in supporting the performance of networks~\cite{8789667}.
There is a great variety of approaches that help experts to satisfy network requirements such as Quality of Service (QoS), network security and resource consumption. 
Starting from the tools that record network data~\cite{joseph2006tcpdump, chappell2010wireshark}, solutions that help an expert to filter the specific data interactively~\cite{10.1007/978-981-10-8944-2_122, 9161633} and finally, automatic systems that detect anomalies or classify the captured network data~\cite{DBLP:journals/comcom/AbbasiST21}.
The use of deep learning (DL) models beneficially eliminated the need to manually search for features in the network data for classification.
DL models learn features from the given raw data and approaches that use different architectures of DL models outperformed all former proposed classifiers~\cite{comp, 8789667}.
The severe drawback of these models is that they are very complex, which is caused by highly complex nonlinear functions that are not interpretable. 
This is why they are called black-box models.
Although these models reach a high accuracy but are not understood by experts which can lead to limited use~\cite{DBLP:journals/corr/RibeiroSG16}.
The reasons for this are low trust and bad interpretability of the model's decisions.
Therefore, the explainability of the decisions of a model is essential, as shown by multiple studies~\cite{xaisurvey, DBLP:journals/corr/abs-2006-11371, chatzimparmpas2020state}.
Besides trust, another important aspect is the evaluation of the model by the expert during the development phase.
Explainable decisions can help to find and correct errors of the model and to improve it, as a model can make false decisions based on misleading data features~\cite{42503}.
For these reasons, it is important to provide an explanation of predictions to an expert in addition to the classification.

In this paper we combine classification of network data, adopt an explanation technique for the classification model and integrate it into an
interactive web-based interface.
Our system represents an interface between experts, algorithms, and data, and therefore plays an important role in fostering trust in classification methods~\cite{9646526}.
Moreover, the expert might extract new insights from the data based on the classification and its explanation. 
This also allows a better evaluation of the model's performance.
We apply a spatial domain attention mechanism, named Class Activation Map (CAM)~\cite{cam}, into a convolutional neural network (CNN) model to classify packets sent over a network and explain the models prediction.
While this method originates from the computer vision field, we adopt it to PCAP data. 
Finally, we present the application class predictions and CAMs to the expert for further investigation.
Our main contributions are:
\begin{itemize}
    \item A web-based visual interface that classifies PCAPs.
    \item Investigation of the most impactful bytes in a PCAP file for a predicted application class. 
    \item Acquisition of new knowledge about PCAP classification with CNN based on the explainability of the classification.
\end{itemize}

The paper is structured as follows. In Section 2, we discuss the related
work on network traffic monitoring and analysis, classification of network files and interpretable DL methods.
Section 3 describes the users, data, tasks and the design requirements. 
Our explainable AI system and its visualization are presented in Section 4. 
In Section 5, we demonstrate a usage scenario on a benchmark dataset.
Section 6 explains our evaluation methodologies and the results. 
Finally, Section 7 concludes our paper with discussion and future work.

\section{Related Work}
We group the related work to our approach into three categories.
First, network traffic monitoring and analysis is discussed.
Then we look at research for classification of network data.
Interpretable DL methods finish the related work section.

\subsection{Network Traffic Monitoring and Analysis}

The field of NTMA has been proven to be a significant topic and has gained attention in the past years. As many networks of different architectures and services have to be monitored regularly to maintain their performance and to fulfill various requirements such as Quality of Service (QoS), traffic security and network data balance.
Different NTMA techniques have been proposed recently in academia as well as in industry~\cite{8789667}.
NTMA techniques can be categorized into two groups: active and passive methods \cite{6863129, DBLP:journals/comcom/AbbasiST21}.
In active methods, test traffic is generated based on planned samples and injected into a network to learn about the condition of the network.
Then various performance metrics are used to measure such characteristics as packet loss ratio and latency in the network in real-time.
Passive methods do not generate artificial traffic but focus on monitoring and the analysis of actual network traffic, particularly in post-event situations.
In such scenario, the experts have to capture the data sent over a network, called packet captures (PCAPs), using packet sniffing tools \cite{chappell2010wireshark, joseph2006tcpdump} and apply various proposed approaches of NTMA such as network traffic classification, traffic prediction, fault management and network security, to satisfy the requirements for network management, troubleshooting and performance \cite{DBLP:journals/comcom/AbbasiST21, 8789667}.
The proposed approaches cover different research areas.
Some approaches are based on information visualization and visual analytics, which help the experts to analyze the data and find complex patterns and abnormal behavior in networks \cite{Guimaraes2016ASO}.
As the collected data grows very fast in size, these tools allow to filter the data for the most relevant parts through interaction by the experts. 
Ulmer et al.~\cite{9161633} proposed NetCapVis, that is a web-based progressive visual analytics system where the user can upload PCAP files and interact with the data and reduce it to a subset based on such attributes as IP addresses, ports, time etc.
Our approach is based on this interface, which is complemented by automatic classification with a neural network and the visualization of the CAM to support the explainability of the predictions.

\subsection{Classification of Network Files}

The main goal of traffic classification is to identify and name different groups of packets using the information available at the network level.
Identification of a packet class can be done by predefined rules.
One of the oldest method of classifying a packet is to query its port number and service name in the register of the Internet Assigned Numbers Authority (IANA) \cite{iana}.
Port-based classifiers label packets by means of the extracted port number from the TCP and UDP headers which is assumed to be associated with a particular application. 
The IANA register functions as a dictionary for this classification.
This type of classifier is often used in firewalls and access control lists due to the fast port extraction from packets \cite{rulebasedclass, 5061972}.
However, these methods have weaknesses, some applications can use dynamic port assignment, hide themselves behind a port numbers already assigned to trusted applications and thus bypass a certain prediction of a class \cite{10.1109/SURV.2009.090304}.
In addition, the number of applications is rapidly changing and growing. 
For all these reasons, more advanced approaches for traffic classification are required to classify modern network traffic.
There are classification techniques based on payload inspection, also known as deep packet inspection (DPI). 
These techniques are built on the analysis of information contained in the payload of packets at the application layer and exploit both packet header and payload for application classification. 
They use predefined patterns starting from the protocol and port numbers, then regular expressions as unique patterns for each protocol and scanning the payload looking for a specific indication, which might refer to a specific application. 
nDPI is one of such open-source tools \cite{npdi}.
Alcock et al.~\cite{payload1} compared the three leading DPI tools L7 Filter, libprotoident and tstat.
In their work it was shown that L7 performs poorly, although it has been used in the research community as ground truth for the traffic classification. 
DPI and libprotoident showed significant better performance.  
The advantage of libprotoident was that it used a minimal amount of payload data.
With constantly changing applications it is difficult to keep a high prediction accuracy, as many patterns need to be regularly updated.
For this reason, it becomes very challenging to maintain the recognizable patterns in the payload and find patterns for new applications.

In recent years DL algorithms have become popular technique in many fields.
In network analysis, DL bypasses the problem of searching for  patterns in the headers and payloads manually by the experts. 
The DL models extract hidden knowledge on large datasets in the training phase.
The first proposed paper with a DL approach was published by Wang~\cite{WinNT}.
In their paper, an Artificial Neural Network (ANN) was applied together with a Stacked Autoencoder (SAE). 
A stacked autoencoder has multiple hidden layers that form a bottleneck in the middle. 
This means that the hidden layers first become smaller and then increase in dimension again.
An autoencoder is a special type of neural network that is trained to reconstruct its input into its output. 
After training, the SAE was extended with a further layer and finetuned for corresponding multiclass classification. 
In this work, SAE usually performed better than the ANN network.
Lotfollahi et al.~\cite{loftani} also compared a SAE model with a 1D Convolutional Neural Network (CNN), called Deep Packet, for classifying encrypted traffic.
CNN is an architecture of neural networks that contains convolutional hidden layers which include a set of independent multiple filters (also called kernels) that perform convolution operations.
The authors used the UNB ISCX VPN-nonVPN dataset to evaluate the performance of the presented method. 
Deep Packet outperformed all previous approaches on this dataset, including two classical ML algorithms, namely k-NN and C4.5.
Their evaluation showed that 1D-CNN outperformed the SAE.
Montieri et al.~\cite{comp} compared models of different DL architectures and also showed that the 1D-CNN performs best. 
The CNN is designed to automatically and adaptively learn spatial dependencies in data and since a packet is considered as a one dimensional sequence, a 1D-CNNs fits best.
These reasons supported the decision to use a 1D-CNN model in our work.
Another reason to use this model is the possibility to extend it with a property to interpret the predictions of the model.

\subsection{Interpretable Deep Learning Methods}
 
DL has become mainstream due to extremely good performance, but it is also known that these models are very complex and difficult to interpret. DL models are often used as black-box models.
Interpretability can make potential properties easier to identify with the help of expertise that can be exploited for further purposes. 
It might help experts determine the root cause and find an appropriate solution.
Interpretability does not improve the performance of a model, however, it is an important part of formulating a highly reliable and trustworthy system \cite{9646526, chatzimparmpas2020state}. 
Moreover, to build trust in DL, interpretability is required and this is an essential factor in many areas.
One of the approaches to make DL models interpretable is to visualize which elements from a sample a hidden neuron is most sensitive to. 
A method proposed by Zhou~\cite{cam} called Class Activation Map (CAM) can be applied to explain the predictions made by a CNN model.
Since a CNN consists of many convolutional layers, the CAM method can exploit the feature map (also called activation map) contained in the filters. 
A filter contains  important features for classification. 
Using the Global Average Pooling (GAP) transformation from the last hidden layer, feature maps are transformed into a single number by averaging, thus producing the weights that are then used to create the resulting CAM.
The advantage of CAM is that it is computationally cheap in comparison to other xAI methods such as SHAP and LIME.
Moreover, the CAM is calculated directly on the resulting trained model, in contrast to methods LIME and SHAP that propose an explanation by learning an interpretable approximated model locally around the prediction \cite{DBLP:journals/corr/RibeiroSG16, lundberg2017unified}. DL models are more complex and the interpretability analysis of the simpler approximated models may not be consistent with the original model.

To the best of our knowledge, only one approach based on CAM has been published by Liu\cite{pcapcam} in network analysis research. 
This work, however, only covered the classification of payload of three datasets labeled with different attack labels.
We, on the contrary, classify applications of encrypted data traffic and provide the explanation of the classification as an interactive visual web-based interface. 

\section{Data-User-Task}
In this section, we first describe the structure of PCAP files. 
We introduce our targeted user groups and their tasks when working with PCAPs analysis to derive the requirements for our approach.

\subsection{Data} \label{data}

Initially, the tcpdump program was developed to capture and decompose network traces~\cite{joseph2006tcpdump}. 
Later, a library was gathered to capture traffic using low-level mechanisms in various operating systems and to read and write network traces, named libpcap~\cite{gharris-opsawg-pcap-00}.
The PCAP file format is a binary format that supports timestamps with nanosecond precision. 
The PCAP files have a general structure, starting with the global header, followed by zero or more packet records.
The global header is found only once at the beginning and has a fixed size of 24 bytes.
This header contains information about the capture characteristics, whether the packet records are saved according to the little endian or big endian depending on the capturing machine, then the time when this file was captured, the local timezone, the accuracy of time stamps in the capture, the number of bytes of packet data actually captured and saved in the file.
A packet record is a container that includes the packets coming from the network.
Each packet record contains packet header and the payload.
The packet header has a length of 16 bytes and includes such information as timestamp and number of bytes of packet saved in file.
The actual packet data immediately follows the packet header as a binary large object.
Generally, the network data has been captured at the data-link layer.
So it contains the Ethernet header providing information about the physical connection of the source and destination Media Access Control address (MAC). 
Then follows the  Internet Protocol (IP) with today's version 4 or 6, where the User Datagram Protocol (UDP) and Transmission Control Protocol (TCP) packets are encapsulated with varying header lengths. 
The TCP packets typically consist of a header of 20 bytes length while UDP packets have an eight byte header.
The nested structure results in one header for each layer at the start of a packet followed by the payload which carries the data.

\subsection{Users} \label{users}

Nowadays, many devices have a network connection and there are multiple ways to capture raw network traffic between the machines.
This means that this approach can be used by a wide range of users, who should have knowledge about network to effectively analyze this data.
That is why our approach is directed at network and cyber analysts.
These experts are interested in network troubleshooting, analysis and finding the cause of anomalies, vulnerabilities and other suspicious traffic in network.
The other group are ML experts and developers, who design classification models based on raw data from the network and provide them to the domain experts. 
In this context, the expert users of our approach are:
\begin{itemize}
\setlength{\itemsep}{0pt} 
    \item Network administrators
    \item Malware analysts
    \item Cybersecurity professionals
    \item ML experts and developers
\end{itemize}
The network experts can benefit most from the proposed visualization interface to analyze network data and gain deeper insights from classifications and the visualization of CAMs to explain the predictions.
By using our tool, the classification model can be evaluated by network experts and through their feedback, ML experts can improve the correctness of the classification model which is also an important aspect of trust~\cite{chatzimparmpas2020state}.

\subsection{Tasks} \label{tasks}

Rapid growth of network traffic requires the appropriate management of network resources.
It is often the case that networks are asymmetrically structured, since the download stream is typically larger than the upload stream.
Therefore, recognizing different types of network applications utilizing the network has become important.   
Consequently, accurate traffic classification gained importance and has become one of the requirements for advanced network management tasks such as adequate resource allocation.
Traffic classification also identifies user behavior and predicts traffic categories, which supports network management. 
ML models provide strong performance in the classification task~\cite{loftani, comp}.
However, a model should not be used as a black-box model but should also provide important keys that help cybersecurity experts make decisions and extract security rules.
With our approach we are not only going to classify PCAPs but also explain the classification by letting the user interactively inspect the prediction scores for each byte.
This allows to extract new knowledge from the data to make appropriate decisions for management and security tasks.
Moreover, by using the model and classification explanation the network experts can provide feedback which the model can be evaluated with.
In case of an identified potential for improvement, the ML experts can perform modifications to the model and data features, e.g. such as masking the IP addresses in the PCAPs, to improve the classification model performance and correctness.
Moreover with our explainable AI approach we strengthen the trust of the user in automatic classification system as the model decisions are explained by salient bytes with different impacting values which provides semantic information of the specific class affiliation. 
To build this trust in the system it is necessary for the user to understand why certain packets are classified as a specific class.
Based on the related work and our previous experience of NetCapVis\cite{9161633}, we identified the following tasks:
\begin{Task}
\setlength{\itemsep}{0pt} 
    \item View classified PCAP data and examine the explanation of the prediction.
    \item Locate relevant features from impacting bytes in PCAP for specific applications. 
    \item  Compare differences in patterns of classified PCAPs for the same and distinct application classes.  
    \item Evaluate strengths and weaknesses of the AI model based on the predictions.
\end{Task}

\subsection{Design Requirements}
Based on the user tasks we derived several requirements for our approach. 
In summary, the requirements are:

\begin{Req}
\setlength{\itemsep}{0pt} 
    \item Show classified PCAPs with its prediction certainty and allow to filter them (\textbf{T1}).
    \item Show explanation for each classification by its impacting feature bytes to extract new knowledge from data and strengthen the trust of the user for automatic classification (\textbf{T2, T3, T4}). 
    \item Show details of a single packet and link the selected information between the CAM visualization, packet details view and plain and hex representation of the selected packet (\textbf{T2, T4}).
\end{Req}

\section{Explainable AI System and Visualization}

First, the details of the preprocessing phase and the architecture of the proposed CNN and CAM are discussed.
Then the web-based interactive interface is described, which enables experts to interact with the classified data and analyze it.

\begin{figure*}[ht]  
\centering
  \frame{\includegraphics[width=0.8\textwidth]{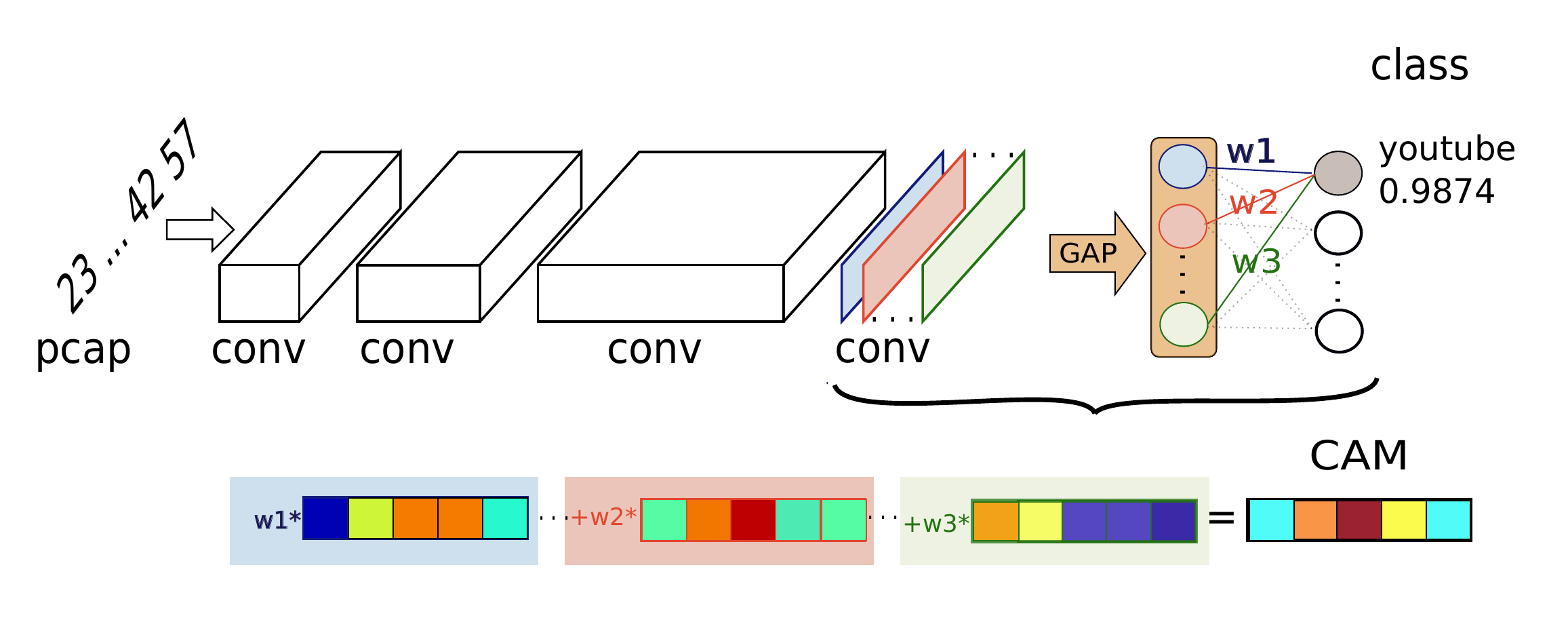}}
  \caption{Illustration of how a CAM is created for a PCAP. A PCAP is shown passing through the CNN model. After this packet has been classified, the CAM is also calculated. The predicted class score is mapped back to the previous convolutional layer to create the CAMs. The resulting CAM is a sum of all convolutional feature maps of the last convolutional layer multiplied by the weights of output layer.}
  \label{modeltrainingview}
\end{figure*}

\subsection{Infrastructure and Technology}

We use a Python backend server with the FastAPI framework~\cite{fastapi}.
Further, we use packet manipulation functions from the Scapy library to preprocess the data~\cite{scapy}. 
Our CNN model and CAM are built and trained with the Keras library~\cite{keras}. 
The frontend application is implemented in JavaScript and uses React.js~\cite{react}.

\subsection{Data Preprocessing}
Processing network packets from a machine learning perspective is not as straightforward as processing images or other fixed size data (Section \ref{data}). 
The network data is not only nested but also variable in size, and there are also different protocols (such as UDP and TCP) on the same network layer or even different versions of the same protocol (such as IPv4 and IPv6).
The real data flow in the network contains some irrelevant packets that are small and do not contain any payload. 
These packets are directly discarded and will not pass through the neural network model, consequently getting directly the label \textit{None}. 
These are packets such as SYN, ACK, or FIN, which are used to establish a connection.
In neural networks, the input layer has a fixed size, so the data samples have to be of the same size. 
For this reason some preprocessing steps are necessary which make all packets the same size by eliminating unnecessary information and padding some headers to the same size.
In the first step, we adapt the TCP and UDP headers to the same length. 
Since the UDP header is smaller, we pad zeros to the end of the UDP segments. 
Then the Ethernet header contains information about the physical connections such as the MAC address, which is essential for forwarding the packets on the network, however it does not conclude any characterization of application from the network layers above.
The last variable segment is the payload of the packets. The payload of each packet is cut to 1480 bytes.
The reason for this is that almost all packets are smaller and only a very small part of the packets contain larger payloads~\cite{loftani}. 
If the payload is smaller it is filled with zeros so that all packets are consistent in size. 
The payload in computer networks is usually limited to 1500 bytes by the Maximum Transmission Unit (MTU).
In the work of Lotfollahi~\cite{loftani} the PCAP packets were investigated in size and they stated that approximately 96\% of packets have a payload length of less than 1480 bytes.
As a result of the preprocessing steps we get a vector of 1500 bytes as the input for our proposed NN. 
Finally, we set the source and destination IP addresses in the PCAP files to zeros so that the model does not erroneously make decisions based on IP addresses.

\subsection{Neural Network Model}

CNNs achieved state-of-the-art results in many areas of pattern recognition, especially in the area of image recognition~\cite{lecun}.
Montieri et al.~\cite{comp} also showed that the CNN performs best for network data.
CNN is an architecture of neural networks that are made of neurons with learnable weights and biases. 
Ordinary neural networks receive a single input vector and transform it through a series of hidden layers whereby each neuron in a layer is connected to all other neurons in the previous layer.
The neurons of a single layer function completely independently of each other.
In contrast, CNNs contain convolutional hidden layers which include of a set  of independent multiple filters (also called kernels) that perform convolution operations.
These filters are initialized randomly and then are learned by the network through backpropagation in the training phase.
These multiple filters slide through the input sample sequence from beginning to end with respect to its dimensions, mapping them sequentially in a feature map (also called activation map) by performing the dot product with the sub-region of the input sample sequence. 
Filters include the parameters such as filter size and stride. 
The use of a filter smaller than the input data is intentional, as it allows the same filter (set of weights) to be multiplied by the input array multiple times at different points on the input.
The filter is purposely applied to each overlapping part of the input.
The stride denotes the number of units by which the filter moves after each operation. 
When performing a standard convolution operation in multiple layers, the data sequence will continuously shrink.
Another issue is that when the filter moves over the original data sequence, it uses the edges of the sample less often whereas the middle region units contribute more.
To solve the problems of shrinking output and data loss, we pad the input with additional borders of zeros called zero padding.
In such manner we can obtain the layer's outputs of the same spatial dimensions as its inputs.
The design of the model, that we apply, consists of convolutional hidden layers of the same size as the input, as illustrated in Figure~\ref{modeltrainingview}.
It is possible to use different dimensions in the convolutional hidden layers, but in the last layer it is necessary to use the same dimension as the input, because the features maps of the last layer are used for the CAM calculation.
The CAM has to be in the same size as input data as it has the function to illustrate impacting values for each byte for the explanation. 
For this reason, we maintained each convolutional hidden layer in the same dimension.
We only changed the number of hidden layers and the number of feature maps in our experiments. 
The CNN is designed to automatically and adaptively learn spatial hierarchies of features. 
Before the final output layer, global average pooling on the convolutional feature maps will be performed.
Global average pooling transforms a feature map into a single number by averaging the numbers in that feature map.
The last layer is a fully connected layer that connects every neuron of resulting applied global average neurons to every neuron in the last layer that produces the output. 
This maps the representation between the input and the desired categorical output.

\subsection{Explainable AI System}

The idea of our explainable approach originated from image classification in the field of computer vision.
The concept was developed for image data showing the impacting regions of an image for the predicted class.
The proposed approach by Lin~\cite{lin2013network} using global average pooling layer after convolutional hidden layers enables the CNN models to have localization ability. 
Localization ability describes the ability to localize the features in the input that was classified by the CNN.
In the case of images, each pixel is considered as input, in our approach, each byte of a packet is considered an input, therefore it is also possible to apply CAM for showing the impacting bytes of a PCAP.
The importance of the regions is calculated by projecting back the weights of the output layer on to the convolutional feature maps.
This technique is called class activation mapping. 
A weighted sum of these values is used to generate the final output (illustrated in Figure~\ref{modeltrainingview}). 
Similarly, we compute a weighted sum of the feature maps of the last convolutional layer to obtain a class activation map.

\subsection{Visualizations and Interactions}

Our work builds on the NetCapVis tool by Ulmer et al~\cite{9161633}.
It is a web-based progressive visual analytics system where experts can upload PCAP files. 
These PCAP files are progressively uploaded and showed in the dashboard. 
It provides different extracted information from the PCAP files. This information is linked to a timeline. 
The user can restrict certain PCAP packets by filtering and then send them to our classification module. 
Our module is responsible for the classification, CAM calculations and visualization of the CAM. 
Our interface is divided in to four visualization elements that we will describe below. 


\subsubsection*{Overview over Packets}

In the upper part there is an overview of the selected packets that have been sent for classification. 
The overview is represented by a list that is illustrated in the Figure~\ref{fig:teaser}(1).
It is possible to manually set the number of packets that will be displayed. 
The columns of the tables are based on the extracted information of the PCAP files and two of them are filled by the model.
The probability and predicted class are returned by the CNN model.
The probability represents the confidence with which the model predicted the corresponding class (R1). 
The predicted class represents the application known by the model, i.e. this class exists in the training dataset on which the model is trained.
Other information to be found in the list are extracted from the headers of the packet. These are: timestamp, source and destination IP address, destination port, which protocol is used, then the size of the payload, by user defined distinction whether the packet is an incoming or outcoming one and finally what type of packet it is.
The user has the possibility to select a packet. 
After having selected a packet further information appears in the packet details view, CAM and plain text and hex representation of the selected packet.

\subsubsection*{Packet Details}

The packet details represent extracted information from the headers such as Ethernet, IP and TCP/UDP which can be seen in Figure~\ref{fig:teaser}(2).
The packet details view is nested with units that can be unfolded.
These details are clickable and are linked to the plain text and hex representation. 
The selected field is highlighted in both visualizations.

\subsubsection*{Class Activation Map}

\begin{figure}[t]  
\centering
  \includegraphics[width=0.47\textwidth]{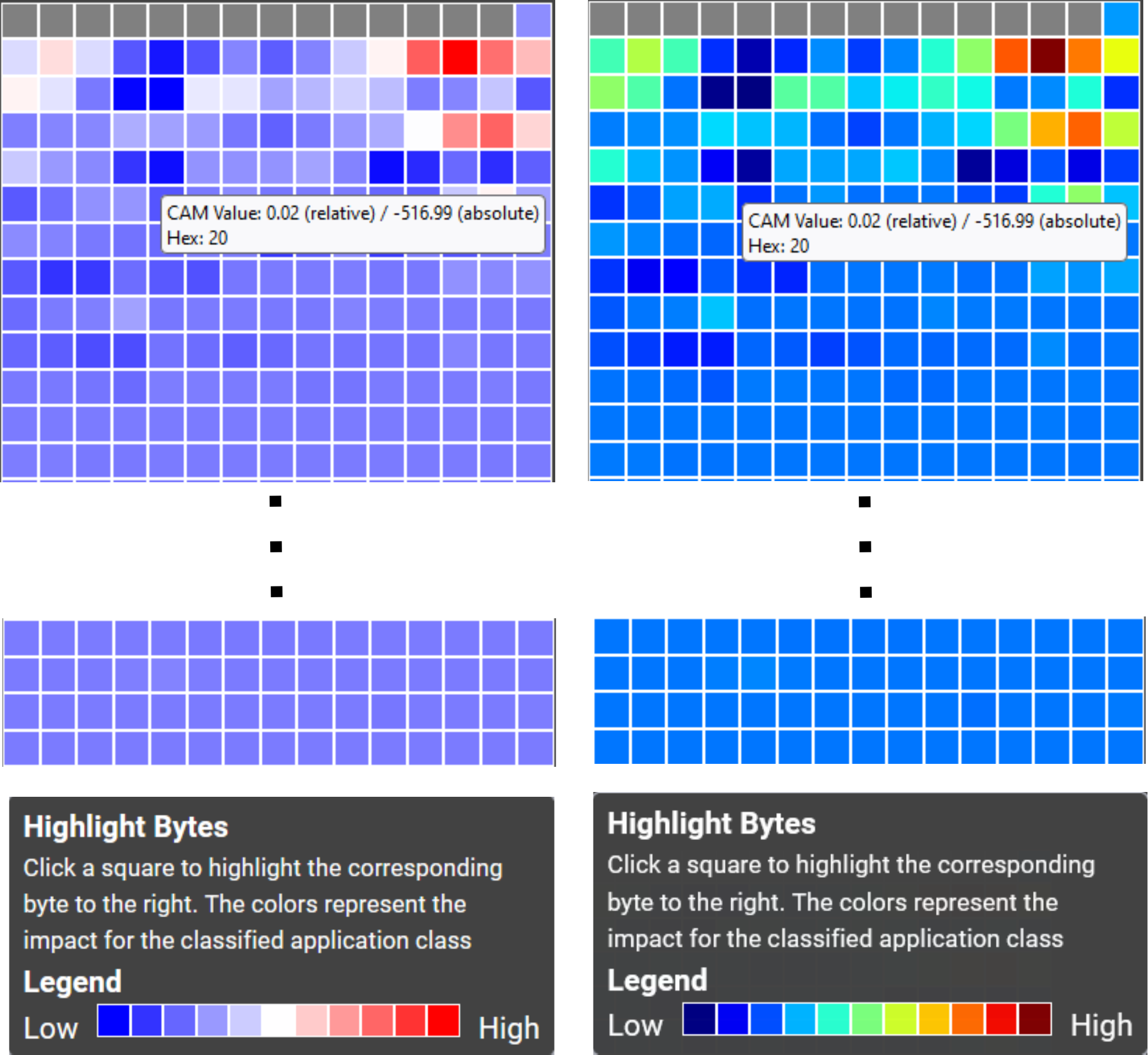}
  \caption{Visualization of an obtained calculated CAM (in two provided colormaps: bwr and jet) in size of 1500 bytes that is repshaped into an image-like vector of a dimension of 15x100. 
  A tooltip illustrates the absolute and relative values of the byte. The colors represent the impact on the corresponding classification. The Ethernet header is grayed out.
because it does not pass through the model.}
  \label{cambeispiel}
\end{figure}

An example of a CAM is illustrated in Figure~\ref{cambeispiel}.
We reshape the obtained calculated CAM in size of 1500 byte into an image-like vector of a dimension of 15x100 as shown in Figure~\ref{cambeispiel}. 
If the payload of the packet is shorter than 1480 bytes and was padded to get this size, it will be shortened back to the original size.
CAM is a heatmap with the maximum size of 15x100 (columns x rows) grids. 
The reason why the CAM is shown as an image-like vector is the alignment with the plain text and hex representation of a PCAP packet. 
The CAM reflects all bytes of a PCAP packet and usually a packet is not represented as one single row but up to a specific byte number and then the new row starts below, therefore, we also used the same logic and represented the CAM in the same way.
The grids in the CAM are linked to the plain text and hex representation of the corresponding packet. 
There is also a tooltip when hovering over a byte in the CAM which illustrates the absolute and relative impacting values of the corresponding byte.
The colors are based on the aforementioned values which illustrate the impacting bytes that led to this class prediction (R2). 
Blue represents absolutely no relevance and the relevance increases with the increase of the red color (R2). 
Since practically all CAM implementations~\cite{cam, 8803474,SUN2022989} use a rainbow colormap to denote their relevance feedback, we have used the same colormap for our technique. 
This is despite the fact that the values of such a quantitative attribute with a minimum and maximum are typically easier to see with a sequential or a divergent colormap. 
The colormap was also mentioned up during the expert interviews and we added a divergent colormap which is selectable in parallel by the user.
The ethernet header is grayed out because it does not pass through the model.

\subsubsection*{Packet Representation}

The plain text and hex representation are shown in Figure~\ref{fig:teaser}(4).
In this visualization, the packets are shown in plain text and hex representations and are linked to all other visualizations. 
The selected elements in the packet details and CAM visualization are highlighted. 
For example, when the IP address is selected in the packet details, the corresponding 4 bytes are highlighted, the same applies to the selected bytes in the CAM (R3).

\section{Usage Scenario} \label{userscena}
This scenario shows a possible usage of our proposed tool, which may help experts to get new insights through the classification and its explanation by the CAMs. 
The expert can strengthen his confidence for the classification as the CAMs indicate the decision making process of the model. 
Another important aspect is the verification of the model. 
By the explainability of the classifications the expert can consider exactly which features have led to this class, thus they can verify the model and in case of wrongly learned features. 
ML expert can then improve the model by the elimination of these features. 
An example of such elimination is masking the IP addresses in PCAP files so that the model does not erroneously make decisions based on IP addresses.

After recording network data, the experts can use the functions of NetCapVis~\cite{9161633} to filter the data by IP addresses, ports and time to relevant parts and send the remaining packets to our classification model. 
Then the remaining packets will be classified by the model and the CAMs are calculated for each packet. 
Afterwards, the user has the possibility to view the predicted application classes for each packet and take a deeper look at the model decisions through the CAM visualization. 
The classification of the packets is performed batchwise in the background.
The predicted classes will be updated batchwise one after the other in the backend's database and later in the overview of all packets in the frontend.
The classified applications with the prediction probability are shown to the expert (T1). 
The expert can focus on a specific class.  
By looking at the CAMs of multiple number of packets, the expert might detect a certain pattern of this class (T2). 
The patterns show the expert the bytes that categorize this class (T3). 
Through the examination of the content in the impacting bytes, the expert can extract new knowledge for this application class and create certain rules in his network such as blocking, prioritization of packets of this class.
If these extracted features appear suspicious for the classification (e.g. IP address), these should be masked and eliminated from the training data and the model should be retrained (T4). 

\begin{figure*}[htb]
\centering
\subfigure[vimeo]{
    \includegraphics[width=.225\textwidth]{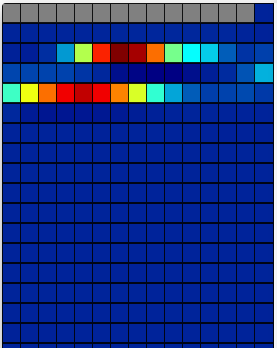}
}
\subfigure[scp]{
    \includegraphics[width=.225\textwidth]{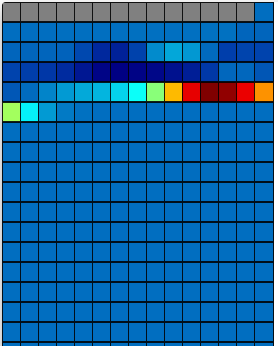}
}
\subfigure[spotify]{
    \includegraphics[width=.225\textwidth]{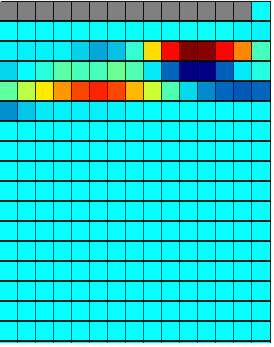}
}
\subfigure[gmail]{
    \includegraphics[width=.225\textwidth]{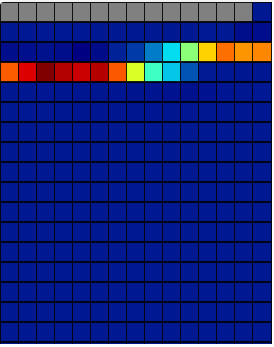}
}
\caption{Representation of four CAMs for vimeo, scp, spority and gmail application classes. Each of these four CAMs represents an average CAM of 100 CAMs of the same class. The purpose of these average CAMs is to crystallize the patterns of corresponding classes (Sect.~\ref{userscena}).}
\label{fig:patterns}
\end{figure*}

We simulated an analysis scenario of a network expert to show how classification of PCAPs and their explainability can be exploited.
We analyzed a predicted application class by taking the CAMs of 100 PCAPs that were classified with a confident probability ($>$90\%) with the same application class and calculated an average CAM from these corresponding 100 CAMs to determine the defining patterns of this class. 
We did that for all classes in the dataset (T3). 
After that, we recorded the impacting bytes of each class which were represented through the corresponding calculated average CAM of this class.
Table \ref{tabpatterns} illustrates significant bytes of each class we extracted. 
We mapped the bytes to the units of the corresponding protocol (T2).  
Figure~\ref{fig:patterns} represents the resulting average CAMs of the classes: vimeo, scp, spotify and gmail. 
The red grids represent the most impacting bytes. 
The expert can directly see which of them influenced the classification and take a look at these bytes (T2). 
Some classes such as ICQ and AIM show classification impact in source and destination port, others illustrate more complicated patterns such as vimeo and scp where also the bytes from the payload are involved.
However, these impacting bytes are at the beginning of the payload, which is probably a not encrypted header of the application.

For this experiment, we used the ISCX VPN-nonVPN dataset, which is well known in network classification studies~\cite{loftani, 8941140, 8004872}.  
Since the dataset is very unbalanced (class FTPS has 7872K samples whereas class AIM has only 5K samples), we adjusted the classes by using random undersampling~\cite{7300975} to approximately the same amount circa 5k per class, similarly as in the work of Lotfollahi et al.\cite{loftani}. 
We created an 1D-CNN model constructed with the following feature map dimensions in the hidden layers: 16, 32, 64, 64, 64, 64, 64, 128, with stride size of 1 and kernel size of 7.
We trained a model 20 epochs on this dataset and achieved a competitive performance F1-Score: 0.98, recall: 0.98, precision: 0.98.

\begin{table*}[htb]
\centering
\caption{Summary of impacting bytes for each application class extracted using the average CAMs for the corresponding class (Sect.~\ref{userscena}). }
\begin{tabular}{lll}
\toprule
\multicolumn{1}{l}{Application} & \multicolumn{1}{l}{Impacting Bytes (Protocol fields)}             & \multicolumn{1}{l}{Protocol} \\ \midrule
aim                               & 1-5 (source/destination ports)                                     & TCP                           \\
email                             & 0-16 (source/destination ports, sequence/acknowledgment numbers)   & TCP                           \\
facebook                          & 8-12 (length, header and data checksum)                            & UDP                           \\
FTPS                              & 9-18 (acknowledgment number, data offset, res, flags, window size) & TCP                           \\
gmail                             & 13-19 (data offset, res, flags, window size)                       & TCP                           \\
hangout                           & 6-7, 12-14 (source/destination ports, acknowledgment number)       & TCP/UDP                       \\
ICQ                               & 4-5 (source/destination ports), payload bytes                      & TCP/UDP                       \\
netflix                           & 8-16 (acknowledgment number, data offset, res, flags)              & TCP                           \\
SCP                               & 21-25 bytes from payload                                      & TCP                           \\
skype                             & 11-16 (acknowledgment number, data offset, res)                    & TCP/UDP                       \\
spotify                           & 6-9 (sequence number), payload bytes                               & TCP                           \\
vimeo                             & 1-6 (source/destination ports), payload bytes                       & TCP                           \\
voipbuster                        & 6-7 (header and data checksum), 12-14 from payload                 & UDP                           \\
youtube                           & 1-4 (source/destination ports)                                     & TCP    \\ \bottomrule

\end{tabular}
\label{tabpatterns}
\end{table*}

\section{Evaluation}

To evaluate our approach, we performed two evaluations: one for the models we used, and one for visual interface.

First, we performed a quantitative evaluation where we tested eight models with different numbers of hidden layers and different kernel sizes. 
In this section, we describe the construction of these models and an overview of the performance by measuring the precision, recall and F1-score.

Second, we individually interviewed experts, in order to evaluate our tool and receive feedback for improvement. 
Three tasks were created for the experts with a real dataset and we let the participants solve these tasks. 
While they were working independently with the data we asked them questions related to the given tasks and the CAM.  
Finally, we asked them to fill out a system usability scale (SUS) form~\cite{brooke1996quick} .

\subsection{Model Evaluation} 

There are datasets of different applications in PCAP format files available as open-source. 
However, some of the records in these datasets are outdated, which means that there are outdated applications that are no longer in use, such as ICQ or AIM chat as can be seen in the ISCX dataset~\cite{datasetispx}. 
For the evaluation of our system with the experts, we focused on the more modern and popular applications. 
To create a representative dataset of real-world traffic, we proceeded in the identical manner as the ISCX VPN-nonVPN dataset was created by ISCX.
Similarly, we created accounts for users Alice and Bob to use services like Youtube, Telegram, etc.
Moreover, we collected the PCAP packets in different IP versions namely in IPv4 and IPv6 respectively for each class.
Table \ref{tab:applicationclasses} represents the number of collected PCAP packets and the corresponding application class. 
PCAP packets of different IP version did not belong in the same class but were separated into two distinct classes.
This means that for each application in Table \ref{tab:applicationclasses} there are two classes.
The PCAP packets were recorded between January 11, 2022 and May 3, 2022 at different times of the day.
We obtained an unbalanced dataset with 3,263,732 samples and 30 classes within.
\begin{table}[ht] \centering
\caption{Our collected dataset of 15 different applications using different IP version. The PCAP packets were recorded between January 11, 2022 and May 3, 2022 at different times of the day. In sum there are 3,263,732 packets in this dataset.}
\begin{tabular*}{0.35\textwidth}[t]{lll}
\toprule
 Application & Size (IPv4) & Size (IPv6) \\ \midrule
 Big Blue Button & 122672 & 150474 \\
 Email & 66564 & 52041 \\
 Facebook Video & 111950 & 62030\\
 FTPS+SFTP & 103701 & 50589 \\
 Google Meet & 90214 & 55059 \\
 Amazon Prime Video & 123329 & 174930 \\
 Reddit & 119221 & 157737 \\
 Telegram Files & 42252 & 41187 \\
 TikTok & 128739 & 138929 \\
 Twitch & 105531 & 138029 \\
 Vimeo & 129652 & 177734 \\
 Youtube & 139248 & 179525 \\
 Zoom & 53029 & 114624 \\
 Instagram & 69872 & 153085\\
 Facebook Feeds & 117894 & 93891 \\ \bottomrule
\end{tabular*}
\label{tab:applicationclasses}
\end{table}

To evaluate the performance of our models, we used the recall (Rc), precision (Pr) and F1-Score (F1) metrics.
The F1-score combines the precision and recall metrics into a single metric.
The F1-score has been designed to work well on imbalanced data~\cite{FERRI200927}. 
These metrics are mathematically described as follows:
\begin{equation}
Pr = \frac{TP}{TP+FP}, 
Rc = \frac{TP}{TP+FN},
F1 = \frac{2*Pr*Rc}{Pr+Rc} 
\end{equation}
Where TP, FP and FN stands for true positive, false positive and false negative, respectively.
We constructed eight models with different numbers of hidden layers. 
The parameters such as stride and padding were the same for these models. The stride was 1 and padding was set so that the dimension of the hidden layers was the same as the dimension of the input.
The models were trained using the categorical cross entropy as loss function~\cite{10.5555/3327546.3327555} and Adam optimizer~\cite{DBLP:journals/corr/KingmaB14}.
Each model was trained and evaluated against the independent test set that was extracted from the dataset (80\% of data was used for training and 20\% for testing).

\begin{itemize}
\setlength{\itemsep}{0pt} 
    \item model1 is constructed with the following feature map dimensions in the hidden layers: 
    16, 32, 64, 128
    \item model2 is constructed with the following feature map dimensions in the hidden layers: 
    16, 32, 64, 64, 64, 64, 64, 128
\end{itemize}
The end of each network is identical, after the last convolutional layer follows GAP and a fully connected layer that has an output of 30 units.
Our intention was to design the final convolutional layer with a sufficient number of feature maps, because these maps produce the resulting CAM. 
These feature maps are weighted and a sufficient number of weighted maps provide a more precise explanation.

We tested whether the performance of the models with different sizes of kernels differs as well as how the number of hidden layers affects the performance.
We observed that both networks perform reliably. 
The smaller network (model1) achieves slower solid performance, which might be due to the smaller number of learnable parameters. 
We also observed that very large filters tend to perform a bit poorer. 
The results are represented in the Table \ref{tab:eval}.
\begin{table}[]
\centering
\caption{Evaluation of 8 different models with varying kernel size and amount of hidden layers, trained and evaluated on our dataset (Table \ref{tab:applicationclasses}). The performance is measured in F1-score.}
\resizebox{0.48\textwidth}{!}{\begin{tabular}{|l|l|llllllllll|}
\hline
\multirow{2}{*}{Model}  & \multirow{2}{*}{Kernel size} & \multicolumn{10}{c|}{Epochs}                                                                                                                                                                                                          \\ \cline{3-12} 
                        &                              & \multicolumn{1}{l|}{1} & \multicolumn{1}{l|}{2} & \multicolumn{1}{l|}{3} & \multicolumn{1}{l|}{4} & \multicolumn{1}{l|}{5} & \multicolumn{1}{l|}{6} & \multicolumn{1}{l|}{7} & \multicolumn{1}{l|}{8} & \multicolumn{1}{l|}{9} & 10   \\ \hline
\multirow{4}{*}{model1} & 3                            & 0.87                   & 0.94                   & 0.96                   & 0.96                   & 0.97                   & 0.98                   & 0.98                   & 0.98                   & 0.98                   & 0.98 \\ \cline{2-2}
                        & 5                            & 0.91                   & 0.96                   & 0.97                   & 0.98                   & 0.98                   & 0.98                   & 0.98                   & 0.98                   & 0.98                   & 0.98 \\ \cline{2-2}
                        & 7                            & 0.94                   & 0.97                   & 0.98                   & 0.98                   & 0.98                   & 0.98                   & 0.98                   & 0.98                   & 0.98                   & 0.98 \\ \cline{2-2}
                        & 9                            & 0.95                   & 0.97                   & 0.98                   & 0.98                   & 0.98                   & 0.98                   & 0.98                   & 0.98                   & 0.98                   & 0.98 \\ \hline
\multirow{4}{*}{model2} & 3                            & 0.95                   & 0.97                   & 0.97                   & 0.97                   & 0.97                   & 0.98                   & 0.98                   & 0.98                   & 0.98                   & 0.98 \\ \cline{2-2}
                        & 5                            & 0.96                   & 0.96                   & 0.96                   & 0.96                   & 0.96                   & 0.96                   & 0.96                   & 0.96                   & 0.96                   & 0.96 \\ \cline{2-2}
                        & 7                            & 0.96                   & 0.97                   & 0.96                   & 0.96                   & 0.97                   & 0.98                   & 0.96                   & 0.98                   & 0.96                   & 0.96 \\ \cline{2-2}
                        & 9                            & 0.96                   & 0.95                   & 0.97                   & 0.95                   & 0.96                   & 0.95                   & 0.96                   & 0.95                   & 0.96                   & 0.95 \\ \hline
\end{tabular}}

\label{tab:eval}
\end{table}

\subsection{Expert interviews}

We interviewed three network experts. 
These experts have experience in the field of network analysis from 5 to 20 years, while only two of them had experience of 5 years in the field of machine learning. 
First, we explained the purpose of our web-based interface, then we showed the functionality of our tool, namely how to upload the dataset and view the classification results. 
Then we explained, on which data the model was trained and what applications it can recognize. 
Finally, we explained what the CAM represents and showed how to interact with the elements of the interface. 
Afterwards, the experts were able to use the tool themselves with a real PCAP dataset.
We prepared three tasks for the experts to test whether these tasks could be solved. 
The experts had to determine the solutions on their own without any help.
The tasks include searching for a packet based on certain information such as predicted application class or probability of classification (T1). 
Lastly, the tasks include interacting with the CAM, which provides the expert with an explanation of the classification.  
In the CAM, the experts had to pick out the most significant bytes (T2). 
During this testing phase, we encouraged the experts to express their thoughts and any questions that came up. 
In the questionnaire, the experts had to fill in the solutions of the three aforementioned tasks, then eight questions about the CAM on a Likert-scale~\cite{Robinson2014, south2022effective} and open field questions.  
Finally, we asked them to fill out a SUS questionnaire.
At the end, we had an informal discussion about possible improvements and whether our design requirements were achieved. 
Detailed information about tasks and questionnaire can be found in the sublimated material.

\subsubsection*{Results}

The three tasks were solved correctly by all experts. 
The experts were able to find the correct packets and locate specific details of these packets considering the CAM and the predicted classification.

We asked the experts whether they have already used such classification systems.
They mostly use the well-known Wireshark tool~\cite{chappell2010wireshark} in their analysis.
The experts have not used any tools for PCAP analysis so far that can predict specific applications based on neural networks.
However, they consider this kind of classification very useful for their analysis.
When asked whether the experts consider automatic classification of packets to be useful, they all answered with the highest rating.

We also asked if the trust in an automatic classification exists without explanation. 
The experts have been skeptical and answered that they first have to use a classification system for a while to build trust.

The CAM has not only the task to explain a classification and make it understandable but also to increase the trust for the prediction. 
Responding to this, experts said that it builds trust through explanation, as the CAM allows the details to be analyzed through the most significant bytes and it does not provide a black-box prediction like other systems. 
However, all experts said that they would have to work with the tool for a longer period of time to further strengthen the trust. 

We asked the experts to formulate the purpose of CAMs with their own words.
The experts described it as the identification of relevant bytes or packet positions of a particular application classification. 
With this we found out that they were able to understand the benefit of CAMs after a short introduction and solving three tasks.
Finally, we asked if the CAM might help to obtain new insights from the PCAP files. 
They agreed with this because they sort out irrelevant parts of the packet during the analysis anyway and the CAM can directly indicate which parts of a packet seem important for the application.  

Then we asked if a grid structure representing a heatmap is a suitable visualization. 
The experts said that the heatmap visualization is intuitive and fits within the alignment of plain text and hex representation. One expert suggested to change the color map, which was done afterwards.
We added a divergent color map (also known as ratio, bipolar or double-ended~\cite{Ware12, Moreland2009DivergingCM}) for the CAM visualization. 
The divergent color map allows us to quickly identify whether CAM values are near most impacting, complete irrelevant for the class or something in between.

The open questions and the discussion after the evaluation brought up helpful  suggestions and triggers for further research.
In the following we summarize the improvement feedback.

\subsubsection*{Improvement Feedback}

There were several suggestions to better highlight important information. At the time of the evaluation, the CAM bytes were not linked to the packet detail tree, but only to the plain text and hex representation of the packet. 
Another suggestion was to highlight the selected grid of the CAM with a frame and in addition frame the complete packet segment to which this byte belongs. 
Highlighting the most important bytes for classification in the packet details tree would be very helpful for finding the important segments in a packet.

One of the suggestions was to set up the overview similar to today's Integrated Development Environment (IDE). 
That means that the list of packets are listed on one side with little space usage with the most significant information such as class and probability. 
The rest, using the full height of the screen, is dedicated to CAM and other packet details such as plain text and hex representation as well as extracted information from the packets. 

A further suggestion was to make certain regions such as header, payload in the CAM noticeable.
The lines of the plain text and hex representation should match the CAM grids in height, which increases the readability, since everything is then in the same alignment.
In the plain text and hex representation the bytes should be separated with a space because the experts were used to it from Wireshark~\cite{chappell2010wireshark}.
We asked the experts which applications they consider interesting for analysis. 
They mentioned to us: Apple Cloud Services, SIP, Webex, MS Teams, Whatsapp (web), Bitcoin, Ethereum, 4Chan, Telnet, SSH, home assistance such as Alexa, Google Home, Siri etc. and network traffic that contains advertising and also adult content.

Finally, the SUS questionnaire scored an average rating of 78. 
This implies a good user satisfaction~\cite{miller2009determining}.

\section{Discussion and Future Work}
\subsection{Discussion}
Our approach allows experts to see which semantic parts of a packet influenced the classification. 
We designed our interface as simple as possible so that an expert can quickly understand the system and start directly an analysis.
The focus of the explainability is to increase the confidence in the classifications. 

Another important aspect is that experts can evaluate the classifications. 
First, the experts might extract new insights from the classifications that they can apply for their work in network management as shown in Section~\ref{userscena}.
Second, the experts might evaluate the classifications of the model, which is important for the research community as they can continue to improve their systems on  the basis of these conclusions. 
An important finding from Section~\ref{userscena} is that when classifying encrypted data only five application classes had the impacting bytes from the payload. 
We showed through the calculated average CAMs of 100 CAMs of each application class that the model does not in most cases make the decisions on the payload bytes of the packets but only learns the properties from the headers and only few application classes such as spotify, vimeo, ICQ, SCP, voipbuster showed impacting bytes from the beginning of the payload which may highly likely unencrypted headers of these applications or higher protocols.
With provided explainability of the model decisions through the CAMs it is observable that the model does not make decisions on the encrypted parts of the packets as has been claimed by using a CNN model as a black-box without classification explainability in other research works~\cite{loftani}.
This shows how important explainable model decisions are.
Therefore, it is necessary to apply more rigorous preprocessing to filter out unencrypted header information to research if machine learning approaches are able to classify encrypted data with high accuracy.

Another challenges we encountered was the unbalanced dataset~\cite{datasetispx}.
The model learned the majority class, so we applied downsampling of this dataset. 
For the model training should be taken into account that there are enough samples of each class and that the distribution of the classes is homogeneous.

\subsection{Future Work}
In addition to the evaluation suggestions and our discussed limitations we have the following ideas for future work. \\
We are going to extend the CAM visualization with a threshold and display the top n of the most impactful bytes directly and expand the packet details tree with the segments that these bytes originate from. \\ 
One feature that would be useful is the creation of a report of the entire classification, i.e. to demonstrate to the expert how the distribution of the predicted classes appears in the data and an overview of the impacting bytes of the corresponding application classes and its content.
\\
Based on the predicated application classes, clustering methods could be applied and combined with other extracted features from the data for visualization purposes. 
Searching for the similar packets by filter functions based on the impacting bytes would be useful. 
This will allow the user to filter out all similar packets where the particular bytes have contributed to a specific application class.  
\\
We are interested in an exact evaluation of the detected patterns for specific classes to evaluate the model and to learn which other fields, that may coincidentally correlate with a class, should be hidden besides the IP and MAC addresses.
Since we discovered that all classes had the impacting bytes in headers and only a few classes had some impacting bytes additionally in the payload, we are going to train our models only on the encrypted data, i.e. completely without any known header information.
Currently we create a vector based on the bytes of a PCAP packet, on which a model is trained.
However, some bytes are not independent, for example, the IP address which consists of 4 bytes. 
Our idea is to keep such grouped entities, i.e. to unite the 4 bytes which represent the IP as one feature for the model. 
This way it could be tested if the impacting bytes can be easier to interpret.
Moreover, this would also make it possible to use other algorithms for classification explanation~\cite{DBLP:journals/corr/LundbergL17}.

\section{Conclusion}

In this paper, we presented a visual interactive tool that serves as an interface
between experts, algorithms, and data. 
We combine classification with a CNN model for network data and adopt an explanation technique to interpret predictions. 
We summarized the related work in NTMA, classification of network files and interpretable deep learning methods. 
We defined four user groups and their tasks and then derived three requirements for our system. 
Based on the requirements we implemented our approach, introduced our system and demonstrated a usage scenario with a well known dataset in network analysis. 
We collected our own dataset with currently popular common applications.
We tested our model with our dataset and a well-known dataset from the research community~\cite{datasetispx}.
The performance of our model showed strong results on both datasets.
Then, we interviewed experts and showed that our tool is intuitive to use and raises trust in automatic classification methods.
Through the interviews we gathered valuable feedback that elicit future research directions more clear.
The reflection and discussion on our work lead to new ideas which we want to tackle in future.

\acknowledgments{
This research work has been funded by the German Ministry of Education and Research and the Hessian State Ministry for Higher Education, Research and the Arts within their joint support of the National Research Center for Applied Cybersecurity ATHENE.}

\bibliographystyle{abbrv-doi}

\bibliography{template}

\begin{thebibliography}{10}

\bibitem{WinNT}
The application of deep learning on traffic identification.
\newblock
  \url{https://www.blackhat.com/docs/us-15/materials/us-15-Wang-The-Applications-Of-Deep-Learning-On-Traffic-Identification-wp.pdf}.
\newblock Accessed: 2022-15-06.

\bibitem{iana}
Iana home page.
\newblock http://www.iana.org/, 2005.

\bibitem{DBLP:journals/comcom/AbbasiST21}
M.~Abbasi, A.~Shahraki, and A.~Taherkordi.
\newblock Deep learning for network traffic monitoring and analysis {(NTMA):}
  {A} survey.
\newblock {\em Comput. Commun.}, 170:19--41, 2021. doi: {{%
10\hspace{.1pt}\discretionary{.}{%
}{.}\hspace{.4pt}1016\discretionary{/}{%
}{/}j\hspace{.1pt}\discretionary{.}{%
}{.}\hspace{.4pt}comcom\hspace{.1pt}\discretionary{.}{%
}{.}\hspace{.4pt}2021\hspace{.1pt}\discretionary{.}{%
}{.}\hspace{.4pt}01\hspace{.1pt}\discretionary{.}{%
}{.}\hspace{.4pt}021}}


\bibitem{comp}
G.~Aceto, D.~Ciuonzo, A.~Montieri, and A.~Pescapé.
\newblock Mobile encrypted traffic classification using deep learning:
  Experimental evaluation, lessons learned, and challenges.
\newblock {\em IEEE Transactions on Network and Service Management},
  16(2):445--458, 2019. doi: {{%
10\hspace{.1pt}\discretionary{.}{%
}{.}\hspace{.4pt}1109\discretionary{/}{%
}{/}TNSM\hspace{.1pt}\discretionary{.}{%
}{.}\hspace{.4pt}2019\hspace{.1pt}\discretionary{.}{%
}{.}\hspace{.4pt}2899085}}


\bibitem{payload1}
S.~Alcock and R.~Nelson.
\newblock Measuring the accuracy of open-source payload-based traffic
  classifiers using popular internet applications.
\newblock pp. 956--963, 10 2013. doi: {{%
10\hspace{.1pt}\discretionary{.}{%
}{.}\hspace{.4pt}1109\discretionary{/}{%
}{/}LCNW\hspace{.1pt}\discretionary{.}{%
}{.}\hspace{.4pt}2013\hspace{.1pt}\discretionary{.}{%
}{.}\hspace{.4pt}6758538}}


\bibitem{9646526}
E.~Beauxis-Aussalet, M.~Behrisch, R.~Borgo, D.~H. Chau, C.~Collins, D.~Ebert,
  M.~El-Assady, A.~Endert, D.~A. Keim, J.~Kohlhammer, D.~Oelke, J.~Peltonen,
  M.~Riveiro, T.~Schreck, H.~Strobelt, and J.~J. van Wijk.
\newblock The role of interactive visualization in fostering trust in ai.
\newblock {\em IEEE Computer Graphics and Applications}, 41(6):7--12, 2021.
  doi: {{%
10\hspace{.1pt}\discretionary{.}{%
}{.}\hspace{.4pt}1109\discretionary{/}{%
}{/}MCG\hspace{.1pt}\discretionary{.}{%
}{.}\hspace{.4pt}2021\hspace{.1pt}\discretionary{.}{%
}{.}\hspace{.4pt}3107875}}


\bibitem{scapy}
P.~Biondi and the Scapy~community.
\newblock Scapy project.
\newblock \url{https://scapy.net/}, 2021.

\bibitem{brooke1996quick}
J.~Brooke.
\newblock {\em "SUS-A quick and dirty usability scale." Usability evaluation in
  industry}.
\newblock CRC Press, June 1996.
\newblock ISBN: 9780748404605.

\bibitem{10.1109/SURV.2009.090304}
A.~Callado, C.~Kamienski, G.~Szabo, B.~Gero, J.~Kelner, S.~Fernandes, and
  D.~Sadok.
\newblock A survey on internet traffic identification.
\newblock {\em Commun. Surveys Tuts.}, 11(3):37–52, jul 2009. doi: {{%
10\hspace{.1pt}\discretionary{.}{%
}{.}\hspace{.4pt}1109\discretionary{/}{%
}{/}SURV\hspace{.1pt}\discretionary{.}{%
}{.}\hspace{.4pt}2009\hspace{.1pt}\discretionary{.}{%
}{.}\hspace{.4pt}090304}}


\bibitem{chappell2010wireshark}
L.~Chappell and G.~Combs.
\newblock {\em Wireshark network analysis: the official Wireshark certified
  network analyst study guide}.
\newblock Protocol Analysis Institute, Chappell University, 2010.

\bibitem{chatzimparmpas2020state}
A.~Chatzimparmpas, R.~M. Martins, I.~Jusufi, K.~Kucher, F.~Rossi, and
  A.~Kerren.
\newblock The state of the art in enhancing trust in machine learning models
  with the use of visualizations.
\newblock In {\em Computer Graphics Forum}, vol.~39, pp. 713--756. Wiley Online
  Library, 2020.

\bibitem{keras}
E.~A. Chollet~F.
\newblock Keras library.
\newblock \url{https://keras.io}, 2015.

\bibitem{DBLP:journals/corr/abs-2006-11371}
A.~Das and P.~Rad.
\newblock Opportunities and challenges in explainable artificial intelligence
  {(XAI):} {A} survey.
\newblock {\em CoRR}, abs/2006.11371, 2020.

\bibitem{npdi}
L.~Deri, M.~Martinelli, T.~Bujlow, and A.~Cardigliano.
\newblock ndpi: Open-source high-speed deep packet inspection.
\newblock In {\em 2014 International Wireless Communications and Mobile
  Computing Conference (IWCMC)}, pp. 617--622, 2014. doi: {{%
10\hspace{.1pt}\discretionary{.}{%
}{.}\hspace{.4pt}1109\discretionary{/}{%
}{/}IWCMC\hspace{.1pt}\discretionary{.}{%
}{.}\hspace{.4pt}2014\hspace{.1pt}\discretionary{.}{%
}{.}\hspace{.4pt}6906427}}


\bibitem{8789667}
A.~D’Alconzo, I.~Drago, A.~Morichetta, M.~Mellia, and P.~Casas.
\newblock A survey on big data for network traffic monitoring and analysis.
\newblock {\em IEEE Transactions on Network and Service Management},
  16(3):800--813, 2019. doi: {{%
10\hspace{.1pt}\discretionary{.}{%
}{.}\hspace{.4pt}1109\discretionary{/}{%
}{/}TNSM\hspace{.1pt}\discretionary{.}{%
}{.}\hspace{.4pt}2019\hspace{.1pt}\discretionary{.}{%
}{.}\hspace{.4pt}2933358}}


\bibitem{react}
Facebook.
\newblock React - a javascript library for building user interfaces.
\newblock \url{https://github.com/facebook/react}.

\bibitem{FERRI200927}
C.~Ferri, J.~Hernández-Orallo, and R.~Modroiu.
\newblock An experimental comparison of performance measures for
  classification.
\newblock {\em Pattern Recognition Letters}, 30(1):27--38, 2009. doi: {{%
10\hspace{.1pt}\discretionary{.}{%
}{.}\hspace{.4pt}1016\discretionary{/}{%
}{/}j\hspace{.1pt}\discretionary{.}{%
}{.}\hspace{.4pt}patrec\hspace{.1pt}\discretionary{.}{%
}{.}\hspace{.4pt}2008\hspace{.1pt}\discretionary{.}{%
}{.}\hspace{.4pt}08\hspace{.1pt}\discretionary{.}{%
}{.}\hspace{.4pt}010}}


\bibitem{datasetispx}
C.~I. for Cybersecurity.
\newblock Vpn-nonvpn dataset.
\newblock \url{https://www.unb.ca/cic/datasets/vpn.html}, 2015.

\bibitem{xaisurvey}
R.~Guidotti, A.~Monreale, F.~Turini, D.~Pedreschi, and F.~Giannotti.
\newblock A survey of methods for explaining black box models.
\newblock {\em CoRR}, abs/1802.01933, 2018.

\bibitem{Guimaraes2016ASO}
V.~T. Guimaraes, C.~M. D.~S. Freitas, R.~Sadre, L.~Tarouco, and L.~Z.
  Granville.
\newblock A survey on information visualization for network and service
  management.
\newblock {\em IEEE Communications Surveys \& Tutorials}, 18:285--323, 2016.

\bibitem{gharris-opsawg-pcap-00}
G.~Harris and M.~Richardson.
\newblock {PCAP Capture File Format}.
\newblock Internet-Draft draft-gharris-opsawg-pcap-00, Internet Engineering
  Task Force.
\newblock Work in Progress.

\bibitem{8941140}
A.~S. Iliyasu and H.~Deng.
\newblock Semi-supervised encrypted traffic classification with deep
  convolutional generative adversarial networks.
\newblock {\em IEEE Access}, 8:118--126, 2020. doi: {{%
10\hspace{.1pt}\discretionary{.}{%
}{.}\hspace{.4pt}1109\discretionary{/}{%
}{/}ACCESS\hspace{.1pt}\discretionary{.}{%
}{.}\hspace{.4pt}2019\hspace{.1pt}\discretionary{.}{%
}{.}\hspace{.4pt}2962106}}


\bibitem{joseph2006tcpdump}
D.~A. Joseph, V.~Paxson, and S.~Kim.
\newblock tcpdump tutorial.
\newblock {\em University of California, EE122 Fall}, 2006.

\bibitem{DBLP:journals/corr/KingmaB14}
D.~P. Kingma and J.~Ba.
\newblock Adam: {A} method for stochastic optimization.
\newblock In Y.~Bengio and Y.~LeCun, eds., {\em 3rd International Conference on
  Learning Representations, {ICLR} 2015, San Diego, CA, USA, May 7-9, 2015,
  Conference Track Proceedings}, 2015.

\bibitem{lecun}
Y.~Lecun, L.~Bottou, Y.~Bengio, and P.~Haffner.
\newblock Gradient-based learning applied to document recognition.
\newblock {\em Proceedings of the IEEE}, 86(11):2278--2324, 1998. doi: {{%
10\hspace{.1pt}\discretionary{.}{%
}{.}\hspace{.4pt}1109\discretionary{/}{%
}{/}5\hspace{.1pt}\discretionary{.}{%
}{.}\hspace{.4pt}726791}}


\bibitem{lin2013network}
M.~Lin, Q.~Chen, and S.~Yan.
\newblock Network in network, 2013.
\newblock cite arxiv:1312.4400Comment: 10 pages, 4 figures, for iclr2014.

\bibitem{pcapcam}
H.~Liu, B.~Lang, S.~Chen, and M.~Yuan.
\newblock Interpretable deep learning method for attack detection based on
  spatial domain attention.
\newblock In {\em 2021 IEEE Symposium on Computers and Communications (ISCC)},
  pp. 1--6, 2021. doi: {{%
10\hspace{.1pt}\discretionary{.}{%
}{.}\hspace{.4pt}1109\discretionary{/}{%
}{/}ISCC53001\hspace{.1pt}\discretionary{.}{%
}{.}\hspace{.4pt}2021\hspace{.1pt}\discretionary{.}{%
}{.}\hspace{.4pt}9631532}}


\bibitem{6863129}
J.~Liu, F.~Liu, and N.~Ansari.
\newblock Monitoring and analyzing big traffic data of a large-scale cellular
  network with hadoop.
\newblock {\em IEEE Network}, 28(4):32--39, 2014. doi: {{%
10\hspace{.1pt}\discretionary{.}{%
}{.}\hspace{.4pt}1109\discretionary{/}{%
}{/}MNET\hspace{.1pt}\discretionary{.}{%
}{.}\hspace{.4pt}2014\hspace{.1pt}\discretionary{.}{%
}{.}\hspace{.4pt}6863129}}


\bibitem{loftani}
M.~Lotfollahi, R.~S.~H. Zade, M.~J. Siavoshani, and M.~Saberian.
\newblock Deep packet: A novel approach for encrypted traffic classification
  using deep learning, 2017. doi: {{%
10\hspace{.1pt}\discretionary{.}{%
}{.}\hspace{.4pt}48550\discretionary{/}{%
}{/}ARXIV\hspace{.1pt}\discretionary{.}{%
}{.}\hspace{.4pt}1709\hspace{.1pt}\discretionary{.}{%
}{.}\hspace{.4pt}02656}}


\bibitem{DBLP:journals/corr/LundbergL17}
S.~M. Lundberg and S.~Lee.
\newblock A unified approach to interpreting model predictions.
\newblock {\em CoRR}, abs/1705.07874, 2017.

\bibitem{lundberg2017unified}
S.~M. Lundberg and S.-I. Lee.
\newblock A unified approach to interpreting model predictions.
\newblock {\em Advances in neural information processing systems}, 30, 2017.

\bibitem{rulebasedclass}
Y.~Luo, K.~Xiang, and S.~Li.
\newblock Acceleration of decision tree searching for ip traffic
  classification.
\newblock pp. 40--49, 01 2008. doi: {{%
10\hspace{.1pt}\discretionary{.}{%
}{.}\hspace{.4pt}1145\discretionary{/}{%
}{/}1477942\hspace{.1pt}\discretionary{.}{%
}{.}\hspace{.4pt}1477949}}


\bibitem{miller2009determining}
A.~B. P. K.~J. Miller.
\newblock Determining what individual sus scores mean: Adding an adjective
  rating scale.
\newblock {\em Journal of Usability Studies}, 4(3):114--123, 2009.

\bibitem{Moreland2009DivergingCM}
K.~Moreland.
\newblock Diverging color maps for scientific visualization.
\newblock In {\em ISVC}, 2009.

\bibitem{7300975}
J.~Prusa, T.~M. Khoshgoftaar, D.~J. Dittman, and A.~Napolitano.
\newblock Using random undersampling to alleviate class imbalance on tweet
  sentiment data.
\newblock In {\em 2015 IEEE International Conference on Information Reuse and
  Integration}, pp. 197--202, 2015. doi: {{%
10\hspace{.1pt}\discretionary{.}{%
}{.}\hspace{.4pt}1109\discretionary{/}{%
}{/}IRI\hspace{.1pt}\discretionary{.}{%
}{.}\hspace{.4pt}2015\hspace{.1pt}\discretionary{.}{%
}{.}\hspace{.4pt}39}}


\bibitem{5061972}
Y.~Qi, L.~Xu, B.~Yang, Y.~Xue, and J.~Li.
\newblock Packet classification algorithms: From theory to practice.
\newblock In {\em IEEE INFOCOM 2009}, pp. 648--656, 2009. doi: {{%
10\hspace{.1pt}\discretionary{.}{%
}{.}\hspace{.4pt}1109\discretionary{/}{%
}{/}INFCOM\hspace{.1pt}\discretionary{.}{%
}{.}\hspace{.4pt}2009\hspace{.1pt}\discretionary{.}{%
}{.}\hspace{.4pt}5061972}}


\bibitem{fastapi}
S.~Ramírez.
\newblock Fastapi framework.
\newblock \url{https://fastapi.tiangolo.com/}.

\bibitem{DBLP:journals/corr/RibeiroSG16}
M.~T. Ribeiro, S.~Singh, and C.~Guestrin.
\newblock "why should {I} trust you?": Explaining the predictions of any
  classifier.
\newblock {\em CoRR}, abs/1602.04938, 2016.

\bibitem{Robinson2014}
J.~Robinson.
\newblock {\em Likert Scale}, pp. 3620--3621.
\newblock Springer Netherlands, Dordrecht, 2014. doi: {{%
10\hspace{.1pt}\discretionary{.}{%
}{.}\hspace{.4pt}1007\discretionary{/}{%
}{/}978\discretionary{%
}{-}{-}94\discretionary{%
}{-}{-}007\discretionary{%
}{-}{-}0753\discretionary{%
}{-}{-}5\_1654}}


\bibitem{south2022effective}
L.~South, D.~Saffo, O.~Vitek, C.~Dunne, and M.~Borkin.
\newblock Effective use of likert scales in visualization evaluations: A
  systematic review.
\newblock 2022.

\bibitem{SUN2022989}
S.~Sun, B.~Song, X.~Cai, X.~Du, and M.~Guizani.
\newblock Cama: Class activation mapping disruptive attack for deep neural
  networks.
\newblock {\em Neurocomputing}, 500:989--1002, 2022. doi: {{%
10\hspace{.1pt}\discretionary{.}{%
}{.}\hspace{.4pt}1016\discretionary{/}{%
}{/}j\hspace{.1pt}\discretionary{.}{%
}{.}\hspace{.4pt}neucom\hspace{.1pt}\discretionary{.}{%
}{.}\hspace{.4pt}2022\hspace{.1pt}\discretionary{.}{%
}{.}\hspace{.4pt}05\hspace{.1pt}\discretionary{.}{%
}{.}\hspace{.4pt}065}}


\bibitem{42503}
C.~Szegedy, W.~Zaremba, I.~Sutskever, J.~Bruna, D.~Erhan, I.~Goodfellow, and
  R.~Fergus.
\newblock Intriguing properties of neural networks.
\newblock In {\em International Conference on Learning Representations}, 2014.

\bibitem{8803474}
T.~Tagaris, M.~Sdraka, and A.~Stafylopatis.
\newblock High-resolution class activation mapping.
\newblock In {\em 2019 IEEE International Conference on Image Processing
  (ICIP)}, pp. 4514--4518, 2019. doi: {{%
10\hspace{.1pt}\discretionary{.}{%
}{.}\hspace{.4pt}1109\discretionary{/}{%
}{/}ICIP\hspace{.1pt}\discretionary{.}{%
}{.}\hspace{.4pt}2019\hspace{.1pt}\discretionary{.}{%
}{.}\hspace{.4pt}8803474}}


\bibitem{9161633}
A.~Ulmer, D.~Sessler, and J.~Kohlhammer.
\newblock Netcapvis: Web-based progressive visual analytics for network packet
  captures.
\newblock In {\em 2019 IEEE Symposium on Visualization for Cyber Security
  (VizSec)}, pp. 1--10, 2019. doi: {{%
10\hspace{.1pt}\discretionary{.}{%
}{.}\hspace{.4pt}1109\discretionary{/}{%
}{/}VizSec48167\hspace{.1pt}\discretionary{.}{%
}{.}\hspace{.4pt}2019\hspace{.1pt}\discretionary{.}{%
}{.}\hspace{.4pt}9161633}}


\bibitem{8004872}
W.~Wang, M.~Zhu, J.~Wang, X.~Zeng, and Z.~Yang.
\newblock End-to-end encrypted traffic classification with one-dimensional
  convolution neural networks.
\newblock In {\em 2017 IEEE International Conference on Intelligence and
  Security Informatics (ISI)}, pp. 43--48, 2017. doi: {{%
10\hspace{.1pt}\discretionary{.}{%
}{.}\hspace{.4pt}1109\discretionary{/}{%
}{/}ISI\hspace{.1pt}\discretionary{.}{%
}{.}\hspace{.4pt}2017\hspace{.1pt}\discretionary{.}{%
}{.}\hspace{.4pt}8004872}}


\bibitem{Ware12}
C.~Ware.
\newblock {\em Information Visualization: Perception for Design}.
\newblock Morgan Kaufmann Series in Interactive Technologies. Morgan Kaufmann,
  Amsterdam, 3 ed., 2012.

\bibitem{10.5555/3327546.3327555}
Z.~Zhang and M.~R. Sabuncu.
\newblock Generalized cross entropy loss for training deep neural networks with
  noisy labels.
\newblock In {\em Proceedings of the 32nd International Conference on Neural
  Information Processing Systems}, NIPS'18, p. 8792–8802. Curran Associates
  Inc., Red Hook, NY, USA, 2018.

\bibitem{10.1007/978-981-10-8944-2_122}
H.~Zhao, W.~Tang, X.~Zou, Y.~Wang, and Y.~Zu.
\newblock Analysis of visualization systems for cyber security.
\newblock In S.~Patnaik and V.~Jain, eds., {\em Recent Developments in
  Intelligent Computing, Communication and Devices}, pp. 1051--1061. Springer
  Singapore, Singapore, 2019.

\bibitem{cam}
B.~Zhou, A.~Khosla, A.~Lapedriza, A.~Oliva, and A.~Torralba.
\newblock Learning deep features for discriminative localization.
\newblock In {\em 2016 IEEE Conference on Computer Vision and Pattern
  Recognition (CVPR)}, pp. 2921--2929, 2016. doi: {{%
10\hspace{.1pt}\discretionary{.}{%
}{.}\hspace{.4pt}1109\discretionary{/}{%
}{/}CVPR\hspace{.1pt}\discretionary{.}{%
}{.}\hspace{.4pt}2016\hspace{.1pt}\discretionary{.}{%
}{.}\hspace{.4pt}319}}


\end{thebibliography}
\end{document}